%% file: main.tex
\documentclass{article}

\usepackage{arxiv}

\usepackage[utf8]{inputenc} % allow utf-8 input
\usepackage[T1]{fontenc}    % use 8-bit T1 fonts
\usepackage{hyperref}       % hyperlinks
\usepackage{url}            % simple URL typesetting
\usepackage{booktabs}       % professional-quality tables
\usepackage{amsfonts}       % blackboard math symbols
\usepackage{amsmath}   
\usepackage{nicefrac}       % compact symbols for 1/2, etc.
\usepackage{microtype}      % microtypography
\usepackage{lipsum}		% Can be removed after putting your text content
\usepackage{graphicx}
\usepackage{natbib}
\usepackage{doi}
\usepackage{wrapfig}
\usepackage{color}
\usepackage{tabularx}
\usepackage{multirow} % 用于多行合并
\usepackage{colortbl} % 用于设置表格颜色
\usepackage{xcolor} % 提供更多颜色选项
\usepackage{cleveref}

\title{PP-DocBee2: Improved Baselines with Efficient Data for Multimodal Document Understanding}

\author{
\vspace{1mm}
Kui Huang$^\dag$ \quad Xinrong Chen$^\dag$ \quad Wenyu Lv$^{\dag\ddag}$ \quad 
Jincheng Liao \quad Guanzhong Wang \quad Yi Liu \\
\vspace{1mm}
\small Baidu Inc. \\
\small \href{mailto:huangkui01@baidu.com}{huangkui01@baidu.com} \quad 
\small \href{mailto:huangkui01@baidu.com}{chenxinrong02@baidu.com} \quad 
\small \href{mailto:lvwenyu01@baidu.com}{lvwenyu01@baidu.com} 
}

\renewcommand{\undertitle}{Technical Report}
\renewcommand{\shorttitle}{PP-DocBee2: Improved Baselines with Efficient Data for Multimodal Document Understanding}
\definecolor{brickred}{rgb}{0.8, 0.25, 0.33}
\definecolor{ao(english)}{rgb}{0.0, 0.5, 0.0}
\newcommand{\increase}[1]{{
  \fontsize{8.5pt}{0.5em}\selectfont({\color{ao(english)}{$\uparrow$~\textbf{#1}}})
}}

\begin{document}
\maketitle

\footnotetext{$^\dag$Equal contribution. $^\ddag$ Project lead.}

\begin{abstract}
This report introduces PP-DocBee2, an advanced version of the PP-DocBee, designed to enhance multimodal document understanding. Built on a large multimodal model architecture, PP-DocBee2 addresses the limitations of its predecessor through key technological improvements, including enhanced synthetic data quality, improved visual feature fusion strategy, and optimized inference methodologies.
These enhancements yield an $11.4\%$ performance boost on internal benchmarks for Chinese business documents, and reduce inference latency by $73.0\%$ to the vanilla version.
A key innovation of our work is a data quality optimization strategy for multimodal document tasks. By employing a large-scale multimodal pre-trained model to evaluate data, we apply a novel statistical criterion to filter outliers, ensuring high-quality training data. 
Inspired by insights into underutilized intermediate features in multimodal models, we enhance the ViT representational capacity by decomposing it into layers and applying a novel feature fusion strategy to improve complex reasoning.
The source code and pre-trained model are available at \href{https://github.com/PaddlePaddle/PaddleMIX}{https://github.com/PaddlePaddle/PaddleMIX}.

\end{abstract}

\section{Introduction}
\input{section/intro}

\section{Method}
\input{section/method}

\section{Experiment}
\input{section/experiment}

\section{Conclusion}
In this report, we introduced PP-DocBee2, an advanced multimodal document understanding model that builds upon the foundation laid by its predecessor, PP-DocBee. Through a series of strategic enhancements, including improved synthetic data quality, refined visual feature fusion strategy, and advanced training methodologies, PP-DocBee2 has demonstrated a significant performance boost in understanding complex document elements, achieving an $11.4\%$ improvement on internal benchmarks for Chinese business documents.
Our ablation studies confirmed the effectiveness of the data filtering strategy and the layer selection approach, yielding the best results. Additionally, the latency reduction techniques implemented in PP-DocBee2 achieved a remarkable $73.0\%$ reduction in inference time, maintaining accuracy while significantly improving efficiency.
Overall, PP-DocBee2 presents a powerful tool for enterprises and individuals to manage extensive document information effectively, especially in complex Chinese-language contexts. 
Looking forward, PaddlePaddle will continue to advance intelligent document processing technologies, aiming to make AI a truly powerful assistant for both enterprises and individuals in handling massive volumes of document information.

{
    \bibliographystyle{unsrtnat}
    \bibliography{main}
}

\end{document}

%% file: section/intro.tex
Since the advent of PP-DocBee~\citep{ni2025ppdocbeeimprovingmultimodaldocument, paddlemix2023}, which achieved remarkable performance in multimodal document understanding, we propose PP-DocBee2 to address several limitations observed in its predecessor. Built upon a multimodal large model architecture, PP-DocBee2 incorporates several key technological upgrades, including enhanced synthetic data quality, improved visual feature fusion strategies, and optimized inference methodologies. These advancements significantly improve the model’s ability to comprehend complex document elements.
Compared to its predecessor, PP-DocBee2 demonstrates an approximate $11.4\%$ improvement on internal Chinese business document benchmarks. It not only enhances the precision of textual information extraction but also deepens the understanding of semantic content within images, enabling an end-to-end pipeline from document image input to structured text comprehension output.
PP-DocBee2 exhibits superior adaptability and accuracy across various document question answering (DocQA) scenarios, including financial report analysis, research report interpretation, contract review, product manual recognition, and legal and regulatory information retrieval. It is particularly well-suited for complex document understanding tasks in Chinese-language contexts.

%% file: section/method.tex
\begin{figure}[t]
  \centering
  \includegraphics[width=0.8\textwidth]{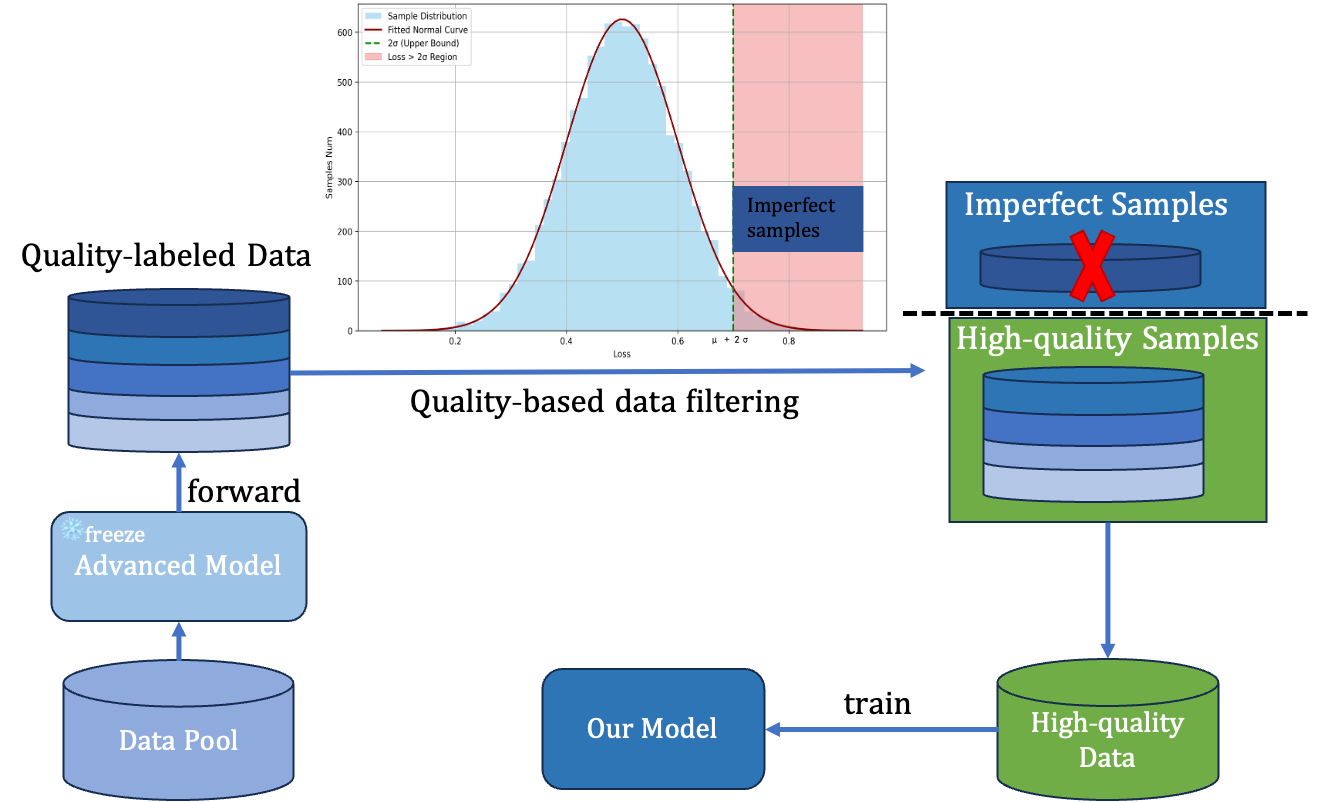} 
  \caption{Loss-based multimodal data sampling strategy.}
  \label{fig:DataSampling}
\end{figure}

We find that accurate and appropriate data play a crucial role in the training of target models. \textbf{Accurate data}, as the name suggests, refers to samples where the semantics across modalities are correct and well-aligned without flaws. \textbf{Appropriate data}, on the other hand, implies that the data present a moderate level of learning difficulty, enabling the model to effectively acquire relevant knowledge during training.
To this end, we propose a data quality optimization strategy tailored for multimodal document understanding tasks, \Cref{fig:DataSampling}. This strategy leverages the discriminative capabilities of strong foundation models to provide cleaner and more consistent data distributions for training lightweight models, thereby enhancing both training efficiency and overall performance.
Specifically, we first employ a powerful large-scale multimodal pre-trained model (e.g., Qwen2.5VL-7B~\citep{bai2025qwen25vltechnicalreport}) as a data evaluator to perform forward inference on the raw document dataset that has not been involved in training. For each sample, we compute the forward cross-entropy loss. Considering that multimodal documents typically contain diverse modalities such as text, images, and layout structures, this loss value serves as a comprehensive indicator reflecting the model's difficulty in understanding semantic alignment across modalities and the task objective.
Subsequently, we model the loss distribution based on the statistical principle based on $2\sigma$ (empirically validated as optimal through ablation studies): samples with loss values exceeding the mean plus two standard deviations ($\mu + 2\sigma$) are discarded as outliers, and only relatively "easy-to-learn" high-quality data are retained. This process effectively filters out potentially noisy or challenging samples in the dataset, thereby enhancing the consistency and representativeness of the training data.

\subsection{Data Pipeline}

In statistics and natural phenomena, the normal distribution is one of the most common data distributions. According to this distribution, we aim to remove samples that are outliers with high difficulty for the model to learn, as these samples may cause disturbances to the well-learned knowledge. Such disturbances can lead the model to produce incorrect answers on questions it would otherwise answer correctly. Formally, for a random variable $X$ following a normal distribution with mean $\mu$ and standard deviation $\sigma$, the probability density function is given by \cref{equ:distri}:

\begin{equation}
\label{equ:distri}
f(x) = \frac{1}{\sqrt{2 \pi \sigma^2}} \exp\left(-\frac{(x-\mu)^2}{2\sigma^2}\right)
\end{equation}

We define outliers as hard negative samples with excessively high loss values (i.e., those located at the upper tail of the normal distribution) in ~\cref{equ:outlier}. To mitigate their negative impact on model training, we follow the empirical 
$\sigma$-based criterion to statically remove a subset of such hard negative examples from the dataset:

\begin{equation}
\label{equ:outlier}
\text{Outlier} \iff x_i > \mu + n*\sigma
\end{equation}

where $1\le n \le 3$. We further fine-tune the model using the filtered dataset, hoping that the model can achieve good generalization from the current data.

\subsection{Architecture}

% \begin{figure}[ht]
%   \centering
%   \includegraphics[width=0.8\textwidth]{pp_arch.png} % 替换为你的图片文件名
%   \caption{Overview of the PP-DocBee2 architecture.}
%   \label{fig:overview}
% \end{figure}

% 我们的模型架构基于qwen-2.5vl 本身架构（``ViT+MLP+LLM''）. movivated by {Rethinking Visual Layer Selection in Multimodal LLMs}, 模型的vit将被区分成shallow, middle, and deep 层，我们将vit的中间特征和模型的最终输出特征进行融合，希望能通过这个融合去增强vit本身的表征能力。

Our model architecture is based on Qwen-VL-2.5~\citep{bai2025qwen25vltechnicalreport}, which adopts a modular ``ViT+MLP+LLM'' structure for visual-language tasks. In this architecture, the Vision Transformer (ViT) serves as the image encoder, followed by an MLP projection layer that aligns visual features into the language model's input space, enabling seamless integration with a large language model (LLM) for downstream tasks such as image captioning, visual question answering, and reasoning.
Motivated by recent insights from \citep{chen2025rethinkingvisuallayerselection}, which highlight the underutilization of intermediate visual features in existing multimodal LLMs, we revisit the internal representation hierarchy of the ViT. Instead of relying solely on the final output from the last transformer layer, we decompose the ViT into three stages: shallow, middle, and deep layers, corresponding to early low-level, mid-level semantic, and high-level abstract representations, respectively.
To fully leverage this rich visual hierarchy, we propose a feature fusion strategy that aggregates representations from the middle layers with those from the final layer. Specifically, we extract token embeddings from one or more middle ViT blocks and combine them with the final visual output before feeding them into the MLP projection head. This design is aimed at enhancing the representational capacity of the visual encoder, enabling the model to retain more fine-grained and multi-scale visual cues that are often lost in deeper transformer layers.
By integrating multi-level visual information, our method encourages the ViT to learn more robust and semantically aligned features that can better support the reasoning capabilities of the LLM. This approach is particularly beneficial for complex visual tasks where both local detail and global context are essential.

%% file: section/experiment.tex
We conduct extensive experiments to demonstrate the effectiveness of our data filtering scheme and visual feature fusion strategy.
By filtering the data, we create a rich and diverse training environment that enhances the model’s generalization capability across a wide range of scenarios and modalities.

\subsection{Implementation Details}
PP-DocBee2 is built upon the Qwen2.5-VL-3B~\citep{bai2025qwen25vltechnicalreport} model. In the supervised fine-tuning (SFT) phase, the visual encoder remains fixed, and only the parameters of the language model are updated. The model is trained for 16,000 steps on a dataset of about 5 million samples with a batch size of 32 on a single machine equipped with 8 NVIDIA A800 GPUs.

% \subsection{Evaluation}

\subsection{Results}
We evaluate the performance of PP-DocBee2 on our internal Chinese business scenario image benchmarks~\citep{ni2025ppdocbeeimprovingmultimodaldocument}, which are divided into four categories: printed text (656 images), tables (358 images), seals (15 images), and charts (167 images).
As shown in Table~\ref{tab:results}, PP-DocBee2-3B model performs well across multiple categories. In particular, for the ``Printed Text'' category, the PP-DocBee2-3B model achieved a leading score of 545, surpassing other models. Additionally, it scored 253 and 47 in the ``Tables'' and ``Charts'' categories respectively, demonstrating strong capabilities in understanding and processing tabular and graphical data. Although the score for ``Seals'' was only 7, the model achieved the highest overall score of 852 among all evaluated models.
These results suggest that PP-DocBee2-3B delivers strong overall accuracy in handling Chinese multimodal data. The model has shown impressive performance in text recognition, table analysis, and interpretation of various visual elements. Consequently, PP-DocBee2-3B stands out as the top-performing model in terms of comprehensive accuracy.

\begin{table}[ht]
    \centering
    \begin{tabular}{c|ccccc}
        \hline
         \multirow{2}{*}{Model} & Overall & Printed Text & Tables & Seals & Charts \\
         & 1196 & 656 & 358 & 15 & 167 \\
         % Models & Overall Score (1196) & Printed Text (656) & Tables (358) & Seals (15) & Charts (167) \\
         \hline
         GPT-4o~\citep{openai2024gpt4o} & 685 & 436 & 198 & 5 & 46 \\
         GLM-4V Flash~\citep{glm2024chatglmfamilylargelanguage} & 547 & 339 & 169 & 5 & 34 \\
         InternVL2.5-2B~\citep{chen2025expandingperformanceboundariesopensource} & 596 & 363 & 182 & 4 & \textbf{47} \\ 
         Qwen2-VL-2B~\citep{wang2024qwen2vlenhancingvisionlanguagemodels} & 680 & 476 & 167 & \textbf{8} & 29 \\
         Qwen2.5-VL-3B~\citep{bai2025qwen25vltechnicalreport} & 789 & 526 & 223 & 6 & 34 \\
         PPDocBee-2B~\citep{ni2025ppdocbeeimprovingmultimodaldocument} & 765 & 517 & 202 & 5 & 41 \\
         \textbf{PPDocBee2-3B} & \textbf{852} & \textbf{545} & \textbf{253} & 7 & \textbf{47} \\
         \hline
    \end{tabular}
    \vspace*{2mm}
    \caption{Performance comparison among various models on internal benchmarks. The internal business Chinese benchmarks cover scenarios such as financial reports, laws and regulations, science and engineering papers, manuals, humanities papers, contracts, and research reports. It is categorized into four main types: Printed Text, Tables, Seals, and Charts.}
    \label{tab:results}
\end{table}

\begin{table}[ht]
    \centering
    \begin{tabular}{c|ccccc}
        \hline
         \multirow{2}{*}{Strategy} & Overall & Printed Text & Tables & Seals & Charts \\
         & 1196 & 656 & 358 & 15 & 167 \\
         % Strategy & Overall Score & Printed Text & Table & Stamp & Chart \\
         \hline
         Vanilla & 794 & 533 & 223 & 6 & 32 \\
         Training on Unfiltered Data & 812 & 522 & 239 & \textbf{7} & 44 \\
         Training on $3\sigma$ Filtered Data & 832\increase{20} & 533 & 250 & \textbf{7} & 42 \\
         Training on $2\sigma$ Filtered Data & \textbf{845}\increase{33} & \textbf{536} & \textbf{256} & \textbf{7} & \textbf{46} \\ 
         Training on $\sigma$ Filtered Data & 836\increase{24} & \textbf{536} & 247 & \textbf{7} & \textbf{46} \\
         \hline
    \end{tabular}
    \vspace*{2mm}
    \caption{Performance comparison among various strategies on internal benchmarks. Sigma corresponds to the Sigma rule in the normal distribution of data.}
    \label{tab:ablation}
\end{table}

\begin{table}[ht]
    \centering
    \begin{tabular}{c|ccccc}
        \hline
         \multirow{2}{*}{Strategy} & Overall & Printed Text & Tables & Seals & Charts \\
         & 1196 & 656 & 358 & 15 & 167 \\
         % Strategy & Overall Score & Printed Text & Table & Stamp & Chart \\
         \hline
         w/o layer selection & 845 & 536 & 256 & \textbf{7} & 46 \\ 
         % \hline
         layer16\_24\_mean & 839 & 531 & 252 & \textbf{7} & 49 \\
         layer\_8\_16\_24\_mean & 833 & 533 & 250 & \textbf{7} & 43 \\
         
         layer\_24 & 840 & 526 & \textbf{257} & \textbf{7} & \textbf{50} \\ 
         layer\_16 & \textbf{852} & \textbf{545} & 253 & \textbf{7} & 47 \\ 
         
         \hline
    \end{tabular}
    \vspace*{2mm}
    \caption{The performance comparison of various strategies on the internal benchmark is based on the $2\sigma$ data filtering scheme, where 'layer' refers to the selection of corresponding layer features for fusion, and 'mean' indicates the averaging of features across multiple layers. }
    \label{tab:ablation_vit}
\end{table}

\subsection{Ablations}
\paragraph{Effectiveness of the Data Filtering Strategy}
To assess the impact of data filtering strategies, we conducted a series of ablation experiments. These experiments compare the performance of the baseline model with models trained under different configurations, as shown in Table~\ref{tab:ablation}.
We divided the data filtering strategy into several Filtered Data schemes. Among them, the $2\sigma$ strategy yielded the best filtering results.
\paragraph{Effectiveness of Layer Selection Strategy} 
As shown in Table~\ref{tab:ablation_vit}, selecting layer $16$ yields better results than both multi-layer fusion and deep layer features, achieving a score of up to 852.
However, inserting features from different layers has varying effects on different tasks. For example, selecting features from middle layers works better for ``Printed Text'', while features from layer $24$ perform better for ``Table''.

\subsection{Latency Reduction}
\textbf{Latency Reduction}. MLLMs usually contains billions of parameters in language model, the inference latency of MLLMs is primarily constrained by the Auto-Regressive next-token prediction paradigm inherent in the language model. We focus on optimizing inference efficiency of language model part while maintaining the superior charts and documentations understanding ability of PP-DocBee2. We use kernel fusion, efficient attention implementation, and token sampling optimization to achieve this\footnote{The efficient inference implementation of PP-DocBee2 is released at \url{https://github.com/PaddlePaddle/PaddleMIX/tree/develop/deploy/ppdocbee2}}. Results are shown in Table~\ref{tab:deploy_inference}, the efficient version of PP-DocBee2 achieves $73.0\%$ reduction in inference time and $48.6\%$ decrease in end-to-end latency, while maintaining comparable accuracy to the vanilla version. Table~\ref{tab:test_condition} summarizes the latency benchmark test conditions for PP-DocBee2.

\begin{table}[htbp]
    \centering
    \begin{tabular}{c|cccc}
        \hline
         Version & Preprocessing Latency~(s) & Inference Latency~(s) & Total Latency~(s) \\
         \hline
        Vanilla & \textbf{0.20} & 1.22 & 1.42 \\
         Efficient Version & 0.40 & \textbf{0.33} & \textbf{0.73} \\
         \hline
    \end{tabular}
    \vspace*{2mm}
    \caption{Average latency comparison of PP-DocBee2 on internal benchmarks, tested on an NVIDIA A800 80GB GPU.}
    \label{tab:deploy_inference}
\end{table}

\begin{table}[htbp]
    \centering
    \begin{tabular}{c|c}
        \hline
         Condition & Value \\
         \hline
         Avg Input Tokens & 2648.7 \\
         Avg Output Tokens & 9.1 \\
         Top P & 0.001 \\
         Top K & 1 \\
         Temperature & 0.1 \\
         Max New Tokens & 512 \\
         \hline
    \end{tabular}
    \vspace*{2mm}
    \caption{Test conditions for PP-DocBee2 latency benchmark.}
    \label{tab:test_condition}
\end{table}